%% file: Main.tex
\documentclass[sigconf]{acmart}
\usepackage{xcolor}
\usepackage{adjustbox}
\usepackage{multicol}
\usepackage{amsmath}
\usepackage{multirow}
\usepackage{graphicx}
\usepackage{pbox}
\AtBeginDocument{%
  \providecommand\BibTeX{{%
    \normalfont B\kern-0.5em{\scshape i\kern-0.25em b}\kern-0.8em\TeX}}}

\copyrightyear{2023} 
\acmYear{2023} 
\setcopyright{acmlicensed}\acmConference[MM '23]{Proceedings of the 31st ACM International Conference on Multimedia}{October 29-November 3, 2023}{Ottawa, ON, Canada}
\acmBooktitle{Proceedings of the 31st ACM International Conference on Multimedia (MM '23), October 29-November 3, 2023, Ottawa, ON, Canada}
\acmPrice{15.00}
\acmDOI{10.1145/3581783.3613836}
\acmISBN{979-8-4007-0108-5/23/10}

\usepackage{enumitem}
\usepackage{multirow}
\usepackage{array}
\usepackage{bm}
\newcolumntype{P}[1]{>{\centering\arraybackslash}p{#1}}

\newcommand\red[1]{\textcolor{red}{#1}}

\begin{document}
\title{PromptMTopic: Unsupervised Multimodal Topic Modeling of Memes using Large Language Models}

\author{Nirmalendu Prakash}
\affiliation{%
  \institution{Singapore University of \\Technology and Design}
  \city{Singapore}
  \country{Singapore}
}
\email{nirmalendu_prakash@sutd.edu.sg}

\author{Han Wang}
\affiliation{%
  \institution{Singapore University of \\Technology and Design}
  \city{Singapore}
  \country{Singapore}
}
\email{han_wang@sutd.edu.sg}

\author{Nguyen Khoi Hoang}
\affiliation{%
  \institution{VinUniversity}
  \city{Hanoi}
  \country{Vietnam}
}
\email{20nguyen.hk@vinuni.edu.vn}

\author{Ming Shan Hee}
\affiliation{%
  \institution{Singapore University of \\Technology and Design}
  \city{Singapore}
  \country{Singapore}
}
\email{mingshan_hee@mymail.sutd.edu.sg}

\author{Roy Ka-Wei Lee}
\affiliation{%
  \institution{Singapore University of \\Technology and Design}
  \city{Singapore}
  \country{Singapore}
}
\email{roy_lee@sutd.edu.sg}

\renewcommand{\shortauthors}{Nirmalendu Prakash, Han Wang, Nguyen Khoi Hoang, Ming Shan Hee, \& Roy Ka-Wei Lee}

\begin{abstract}
The proliferation of social media has given rise to a new form of communication: memes. Memes are multimodal and often contain a combination of text and visual elements that convey meaning, humor, and cultural significance. While meme analysis has been an active area of research, little work has been done on unsupervised multimodal topic modeling of memes, which is important for content moderation, social media analysis, and cultural studies. We propose \textsf{PromptMTopic}, a novel multimodal prompt-based model designed to learn topics from both text and visual modalities by leveraging the language modeling capabilities of large language models. Our model effectively extracts and clusters topics learned from memes, considering the semantic interaction between the text and visual modalities. We evaluate our proposed model through extensive experiments on three real-world meme datasets, which demonstrate its superiority over state-of-the-art topic modeling baselines in learning descriptive topics in memes. Additionally, our qualitative analysis shows that \textsf{PromptMTopic} can identify meaningful and culturally relevant topics from memes. Our work contributes to the understanding of the topics and themes of memes, a crucial form of communication in today's society.\\
\red{\textbf{Disclaimer: This paper contains sensitive content that may be disturbing to some readers.}}
\end{abstract}




\maketitle

\section{Introduction}
\input{Introduction}

\section{Related Work}

\input{Related}

\section{Preliminaries}

\input{Preliminary}

\section{Methodology}
\input{Model}

\section{Experiment}
\input{Experiment}

\section{Conclusion}
\input{Conclusion}


\bibliographystyle{ACM-Reference-Format}
\balance
\bibliography{ref}

\newpage
\appendix
\section{Appendix}
\input{Appendix}

\end{document}

%% file: Introduction.tex
\textbf{Motivation.} The proliferation of social media has given rise to a new form of communication: memes. Memes are multimodal and often contain a combination of text and visual elements that convey meaning, humor, and cultural significance. Understanding the topics and themes of memes is important for a variety of applications, including content moderation, social media analysis, and cultural studies. Although meme analysis has been an active area of research for some time, with studies focusing on various aspects such as sentiment analysis~\cite{hu2018memo}, semantics~\cite{tang2003sem,hanselmann2008sem}, abusive content~\cite{pramanick2021momenta,lippe2020multimodal}, and cultural significance~\cite{cannizzaro2016internet, glitsos2019pepe}, there has been little work done on unsupervised multimodal topic modeling of memes. This represents a significant research gap to address, as memes have become an essential form of communication, and understanding their topics and themes is crucial for various applications.

\begin{figure}[t] 
	\centering
	\includegraphics[scale = 0.6]{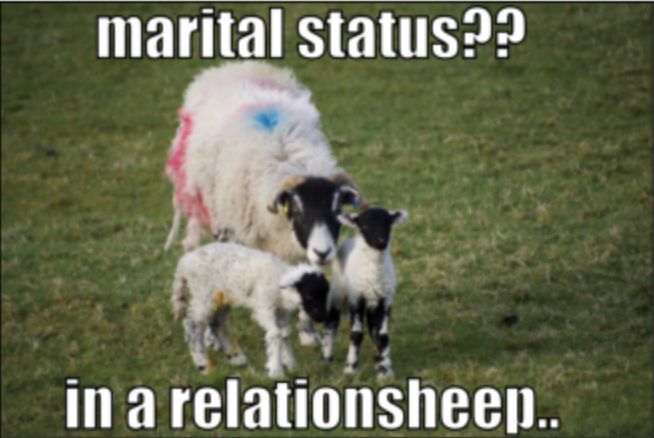} 
	\caption{Example of a multimodal meme from FHM dataset.}
	\label{fig:example}
\end{figure}

A significant challenge in performing unsupervised multimodal topic modeling of memes lies in comprehending the semantic interaction between the text and visual modalities. For instance, in Figure~\ref{fig:example}, if we consider the superimposed text and image separately, they could be associated with topics such as "Marriage" and "Animals" respectively. However, when we consider both modalities together, the appropriate topic should be "Humor" or "Relationship Humor". Hence, it is essential to consider both text and visual elements together to extract meaningful topics from memes. Despite this, most existing topic modeling techniques are predominantly text-based~\cite{vayansky2020topicmodeling} and are not designed to handle multimodal data.



\textbf{Research Objectives.} To address these research gaps, we propose \textsf{PromptMTopic}\footnote{https://github.com/Social-AI-Studio/PromptMTopic}, a novel multimodal prompt-based model designed to learn topics from both text and visual modalities by leveraging the powerful language modeling capabilities of large language models (LLMs). Specifically, \textsf{PromptMTopic} first utilizes a visual language model to extract descriptions of the visual aspect of the meme. These visual captions are then combined with the superimposed text extracted from the memes and used as input to a series of carefully constructed prompts. These prompts guide the LLM in identifying and grouping relevant topics in memes.


\textbf{Contributions.} We summarize our contributions as follows:

\begin{itemize}
\item We propose a novel multimodal prompt-based model to perform topic modeling on memes. To the best of our knowledge, this is the first multimodal topic modeling model that effectively extracts and clusters topics learned from memes.
\item We evaluate our proposed model through extensive experiments on three real-world meme datasets. Our automatic evaluations demonstrate that our proposed model outperforms the state-of-the-art topic modeling baselines in learning descriptive topics in memes.
\item We also perform a qualitative analysis of the learned topics and show that our model can identify meaningful and culturally relevant topics from memes.
\end{itemize}

%% file: Related.tex
\subsection{Multimodal Meme Analysis}

Memes have become an increasingly ubiquitous form of digital communication, representing an impactful means of online socio-political engagement \cite{ross2019internet,chen2022categorizing,nainani2022categorizing} and a fundamental component of cultural narratives \cite{waddock2016foundational}. As such, researchers from various disciplines have attempted to analyze memes to gain insights into the social, cultural, and political issues that concern online communities. Several studies have emphasized the importance of interpreting the combined information from different modalities in meme analysis due to their subtle nature \cite{kiela2020hm, suryawanshi-etal-2020-multimodal,vidgen2019challenges}.

Recognizing the significance of employing a multimodal approach to meme analysis, many studies have released large meme datasets to promote supervised multimodal tasks on memes. For instance, Facebook's Hateful Memes Challenge \cite{kiela2020hateful}, designed to challenge unimodal classifiers, encourages the development of new multimodal solutions to detect hate speech in memes \cite{zhu2020enhance,zhu2022multimodal,hee2022explaining,cao2023prompting,lee2021disentangling,hee2023decoding}. Suryawanshi et al. released the multiOFF dataset and a multimodal offensive content classifier for memes based on this dataset \cite{suryawanshi-etal-2020-multimodal}. Additionally, Memotion \cite{sharma-etal-2020-semeval} promotes the analysis of visuo-lingual metaphors in memes through sentiment classification and emotion recognition. Among notable studies for this task, SESAM \cite{bonheme2020sesam} emphasizes the challenges in combining images and text, showing that alignment-based and fusion-based strategies do not perform as well as using a single modality.

Nevertheless, most of these existing meme studies have centered around performing specific supervised classification tasks. Few studies have explored the underlying topics of memes, and those that have are largely limited to the analysis of the textual modality \cite{vayansky2020topicmodeling}. To address this gap in the literature, we propose an unsupervised multimodal topic modeling approach for memes. Our approach leverages the combined information from both text and image modalities to automatically discover meaningful underlying topics within memes.

\subsection{Topic Modeling}
Topic modeling is a class of unsupervised machine learning techniques that aims to discover the underlying themes within a large corpus of text. Each topic is represented by a list of words most strongly associated with it. Traditional models, such as Latent Dirichlet Allocation (LDA) \cite{blei2003latent} and Non-negative Matrix Factorization (NMF) \cite{fevotte2011algorithms}, treat each document as a bag-of-words (BOW) and assume it to be a mixture of topics. However, this approach neglects the semantic relationships between words. Recent advances in topic modeling, such as Contextualized Topic Models (CTM) \cite{bianchi2020cross} and BERTopic \cite{grootendorst2022bertopic}, incorporate contextual information to provide more nuanced analysis. CTM leverages pre-trained language models such as BERT to capture contextual relationships among words, while BERTopic uses Sentence-BERT (SBERT) \cite{reimers2019sentence} to semantically embed documents into a vector space for comparative purposes.

While unimodal topic modeling is well-studied, research on multimodal topic modeling, which incorporates both image and text modalities, remains limited. Prior studies have often treated visual and textual data as separate entities. For example, tr-mmLDA \cite{putthividhy2010mmlda} learns separate sets of topics for each modality and uses a regression module to predict one set of topics from the other. However, given the inherent interplay between images and text in memes, it is crucial to jointly model both modalities, as illustrated in Figure~\ref{fig:example}. In our study, we propose a framework that first extracts the semantic information of the visual modality in a meme and represents this information as a textual description (i.e., image captions). The textual descriptions are subsequently combined with the superimposed text extracted from the meme to perform topic modeling. By "flattening" the meme into a textual representation, we can apply existing topic modeling methods to learn the topics in memes. Nevertheless, our proposed approach differs from existing topic modeling methods by designing a prompt-based, in-context learning framework that effectively utilizes large language models (LLMs) to learn topics in memes. We theorize that the vast knowledge accumulated in LLMs can potentially enhance the performance of topic modeling, yielding more meaningful topic representation.

%% file: Preliminary.tex
\begin{figure*}[ht!] 
	\centering
	\includegraphics[scale = 0.68 ]{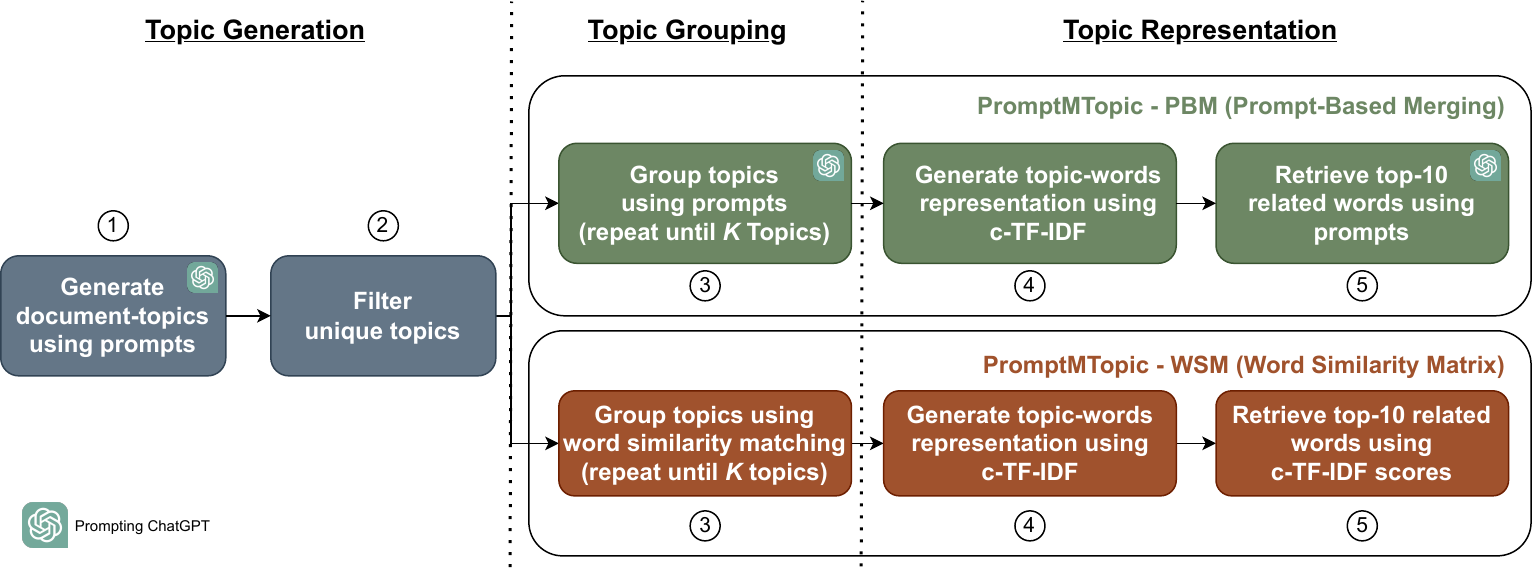} 
	\caption{PromptMTopic Model.}
	\label{fig:framework}
\end{figure*}

\subsection{Problem Definition}
The problem of multimodal meme topic modeling involves identifying common topics within a collection of memes, using both visual and textual data. The problem can be defined as follows: Given a collection of $n$ memes $D = { m_1, m_2, … , m_n }$, with each meme $m = {i, l}$ comprising an underlying image $i$ and superimposed text $l$, the goal is to generate the latent topics and their associated concept probabilities. Traditionally, topic modeling techniques compute a score for each meme-topic pair and topic-word pair through matrix decomposition or the computation of statistical distributions.



\subsection{Feature Extraction}
\label{subsection:feature_extraction}
Existing statistical and embedding-based topic modeling techniques require textual data for matrix decomposition or for the computation of statistical distributions. Therefore, to apply these techniques to memes, it is necessary to (1) extract the superimposed text $l$ and (2) convert the image $i$ into an acceptable textual representation. Superimposed text can be extracted automatically using existing libraries such as EasyOCR or keras-ocr, or by human annotators. Regarding the visual element, a common approach for image-to-text conversion involves representing the image’s semantics with a textual description, or \textit{image captioning}. Hence, we cleaned the superimposed text from the images and input the resulting images into a pre-trained image captioning model. For this task, we used the BLIP-2 model, which has demonstrated its ability to generate high-quality captions for crowdsourced images by leveraging the implicit knowledge in frozen large language models such as Open Pre-trained Transformers \cite{zhang2022opt}. Using BLIP-2, we were able to generate captions that effectively describe the dominant objects or events depicted in the meme's image.

%% file: Model.tex
Our PromptMTopic model is an unsupervised approach that leverages the strong language understanding capabilities of ChatGPT for topic generation. Recent studies have demonstrated that ChatGPT achieves state-of-the-art performance on many natural language tasks \cite{shen2023chatgpt,surameery2023chatgpt,lamichhane2023chatgpt}. Notably, we observed that ChatGPT\footnote{\url{https://api.openai.com/v1/chat/completions}} excels at understanding informal language, including internet jargon and slang. To perform the topic modeling task with ChatGPT, we designed two distinct prompts that extract relevant topics and consolidate a multitude of generated topics into a unique set of themes. Each of these prompts is structured as a multi-turn conversation, where prior turns serve as task demonstrations that guide the model's generation. This process aligns with the concept of \textit{in-context learning}, where large language models are prompted with instructions and/or demonstrations to solve new tasks without additional training \cite{chen2023fs}. This approach enables the model to learn and perform the task efficiently without the need for fine-tuning. Figure \ref{fig:framework} provides an overview of our PromptMTopic model, which consists of two stages: \textit{Topic Generation} and \textit{Topic Collapse}.



\definecolor{question_color}{HTML}{1B9E77}
\definecolor{label_color}{HTML}{D95F02}
\definecolor{explanation_provided_color}{HTML}{7570B3}
\definecolor{explanation_generated_color}{HTML}{FF55A3}
\definecolor{tweet_color}{HTML}{FFC300}

\begin{table}[t!]
\centering
\small
\caption{Example ChatGPT's prompt for \textit{topic generation} (reformatted for visualisation purposes). The prompt input consists of the extracted textual features (in \textcolor{tweet_color}{yellow}), the question (in \textcolor{question_color}{green}), the demonstration answers provided (in \textcolor{explanation_provided_color}{purple}) and the ChatGPT's generated answers (in \textcolor{explanation_generated_color}{pink})}
\begin{tabular}{p{0.90\linewidth}}
\toprule
\# System Instructions \\
\textbf{System}: You are designated as an assistant that identify and extract high-level topics from memes. You should avoid giving specific details and provide unique topics solely. \\ 
\\

\# Demonstration Samples $1 ... N$\\
\textbf{User}: \textcolor{question_color}{Please list the high-level topics in the following meme using its image caption and superimposed text}. \\
Meme's Image Caption: \textcolor{tweet_color}{president obama in front of the white house} \\
Meme's Superimposed Text: \textcolor{tweet_color}{i did not divide the country the republican decision to obstruct every single thing i proposed to help us dig out of the financial crisis they caused divided the country} \\
Topics: \\
\\
\textbf{Assistant}: \textcolor{explanation_provided_color}{['Politics', 'Financial Crisis']}\\
\\

\# New Meme Sample For Topic Generation \\
\textbf{User}: \textcolor{question_color}{Please list the high-level topics in the following meme using its image caption and superimposed text}. \\
Meme's Image Caption: \textcolor{tweet_color}{a man in a car with the caption, when you're in the car and you see in the car} \\
Meme's Superimposed Text: \textcolor{tweet_color}{recruit: why tf gotta wear my id tag and bring my FAD everywhere? BMT spec: IT'S the LAW.} \\
Topics: \\
\\
\textbf{Assistant}: \textcolor{explanation_generated_color}{['Military']}\\
\bottomrule
\end{tabular}
\label{tab:topic_generation_prompt}
\end{table}

\begin{table}[t!]
\centering
\small
\caption{Example ChatGPT's prompt for \textit{topic collapse} (reformatted for visualisation purposes). The prompt input consists of the extracted textual features (in \textcolor{tweet_color}{yellow}), the question (in \textcolor{question_color}{green}), the demonstration answers provided (in \textcolor{explanation_provided_color}{purple}) and the ChatGPT's generated answers (in \textcolor{explanation_generated_color}{pink})}
\begin{tabular}{p{0.90\linewidth}}
\toprule
\# System Instructions \\
\textbf{System}: You are designated as an AI assistant that determine the most appropriate overarching topic from a given list of topics that best corresponds to a current topic. Specifically, you are required to identify the most fitting topic from the provided list that is associated with the given topic, without providing any topics that are not included in the given list of topics. \\ 
\\

\# Demonstration Sample \\
\textbf{User}: \textcolor{tweet_color}{Topics:}\\
\textcolor{tweet_color}{1. Military} \textcolor{tweet_color}{, ... ,} \textcolor{tweet_color}{6. Healthcare} \\ \\
\textcolor{tweet_color}{Current topic : War} \\ \\
\textcolor{question_color}{What is the most suitable topic from the given topics that corresponds to the current topic?}
\\ \\
\textbf{Assistant}: \textcolor{explanation_provided_color}{['Military']}\\
\\

\# New Meme Sample For Topic Generation \\
\textbf{User}: \textcolor{tweet_color}{Topics:} \\ \textcolor{tweet_color}{[$t_1 , t_2, ... , t_K$]}\\ \\
\textcolor{tweet_color}{Current topic : $t_{K+1}$} \\ \\
\textcolor{question_color}{What is the most suitable topic from the given topics that corresponds to the current topic?}
\\ \\
\textbf{Assistant}: \textcolor{explanation_generated_color}{$t \in \{ t_1 , t_2, ... , t_K \} $} \\
\bottomrule
\end{tabular}
\label{tab:topic_grouping_prompt}
\end{table}

\begin{table}[t!]
\centering
\small
\caption{Example ChatGPT's prompt for \textit{topic representation words} (reformatted for visualisation purposes). The prompt input consists of topic and the corresponding top 100 words (in \textcolor{tweet_color}{yellow}), the demonstration answers provided (in \textcolor{explanation_provided_color}{purple}) and the ChatGPT's generated answers (in \textcolor{explanation_generated_color}{pink})}
\label{tab:topic_word_ordering_prompt}
\begin{tabular}{p{0.90\linewidth}}
\toprule
\# System Instructions \\
\textbf{System}: You are designated as an AI assistant that given a list of words and a topic, determines the top 10 words related to the topic, without providing any words that are not included in the above list of words. \\ 
\\

\# Demonstration Sample \\
\textbf{User:} Topic : \textcolor{tweet_color}{'technology'} \\ 
Words : \textcolor{tweet_color}{1. singapore, 2. tracetogether} \textcolor{tweet_color}{, ... ,} \textcolor{tweet_color}{16. message ...} \\ \\
\textcolor{question_color}{Given the above topic, list the top 10 words related to the topic.} \\ \\
\textbf{Assistant}: \textcolor{explanation_provided_color}{['computer', 'robot', 'google', 'iphone', 'airpods', 'internet', 'tracetogether', 'video', 'tiktok', 'message']}
\\ \\

\# New Meme Sample For Topic Generation \\
\textbf{User:} Topic : \textcolor{tweet_color}{$t$} \\ 
Words : \textcolor{tweet_color}{[$w_1 , w_2, ... , w_k$]} \\ \\

\textcolor{question_color}{Given the above topic, list the top 10 words related to the topic.} \\ \\
\textbf{Assistant}: \textcolor{explanation_generated_color}{$[s_1, s_2, ..., s_{10}]$}
\\
\bottomrule
\end{tabular}
\end{table}



\subsection{Topic Generation}
Table \ref{tab:topic_generation_prompt} shows an example of a prompt for topic generation. To guide the behavior of ChatGPT (``assistant''), we meticulously crafted a system message that instructs the assistant to identify and extract high-level topics. Moreover, we instructed the assistant to provide only unique topics and to avoid overly verbose details. We then introduced \textit{N} conversation turns, containing prompt inputs and expected answers, for the model to comprehend and perform the task. For our experiments, we chose \textit{N} = 8 conversation turns as there were noticeably fewer hallucinations with this setting. Finally, we fed the prompt input for the new sample and obtained the topics for individual memes.

However, we observed two scenarios where ChatGPT sometimes failed to generate topics. First, ChatGPT may not comprehend the sample meme due to unclear or incoherent textual inputs. These occurrences are typically because the meme has too few superimposed words, which does not provide enough context for ChatGPT to infer from. Second, the content moderation layer in the ChatGPT API detects inappropriate inputs, such as profanity, and prevents ChatGPT from generating topics. To address these situations, we classified these memes into “Miscellaneous” and “Inappropriate” topic clusters, respectively.


\subsection{Collapsing Overlapping Topics}

While ChatGPT generates multiple high-level topics for individual memes, many of these topics appear similar and overlap. For instance, topics such as “occupation”, “job search”, “labour” and “employment” can be consolidated under the umbrella term “employment”. To address this issue, we propose two approaches for grouping similar topics: \textit{Prompt-Based Matching (PBM)} and \textit{Word Similarity Matching (WSM)}.

 
\textbf{Prompt-Based Matching}. To group similar topics from a collection of memes, we prompt ChatGPT to combine unique topics until a specified number of topics, denoted by $K$, is reached. Initially, we sort the unique topics in descending order based on their respective frequency counts, creating the set $T_{n} = {t_1, t_2, …, t_{n}}$. The rationale behind this is that topics with higher occurrence rates are likely to be more distinctive. Subsequently, we select the first \textit{n}-1 topics from $T_{n}$ to form the set $T_{n-1} = {t_1, t_2, …, t_{n-1}}$. We then prompt ChatGPT to merge the current topic $t_n$ into a topic within the set $T_{n-1}$. If ChatGPT fails to merge $t_n$ with any of the topics in $T_{n-1}$, we incorporate $t_n$ into the “Miscellaneous” topic. We repeat this process until we reduce $T_{n}$ to the set ${t_1, t_2, …, t_K}$. See Table \ref{tab:topic_grouping_prompt} for an example of a topic grouping prompt.

In our experiment, we encountered an issue with some datasets having a large number of unique topics, which exceeded the maximum token length allowed by the ChatGPT API. To overcome this, we used a sliding window of M topics from the sorted unique topic set and performed topic grouping iteratively. If ChatGPT successfully merged a topic into one of the M topics, we merged the topic and restarted the iteration cycle. However, if ChatGPT failed to merge the current topic into any topic in the unique topic list, we assigned the topic to a category named “Miscellaneous”.

\textbf{Word Similarity Matching}. In order to consolidate similar topics into a specified number of unique topics, denoted by $K$, we calculate the similarity between pairs of topics. Specifically, we aggregate the documents belonging to each topic and compute the Class-based Term Frequency - Inverse Document Frequency (c-TF-IDF) word representation:

\begin{equation} \label{eq1}
\begin{split}
    W_{x,c}  = tf_{x,c} \log(1 + \frac{1 + \textit{C}}{1 + df_{x,c}})
\end{split}
\end{equation}

where $tf_{x,c}$ represents the frequency of word $x$ in class $c$, $df_{x,c}$ represents the number of classes in the document set that contain the word $x$, and $C$ represents the total number of classes in the document set. The resulting representation captures the frequency of words in each topic, weighted by their importance across all topics. We then select the top 20 words from the c-TF-IDF representation for each topic. To measure the similarity between topics, we count the number of common words between the top 20 words of each topic pair. If a topic has fewer than 20 words, we normalize the number of common words by the smaller word count between the two topics. This way, we identify topics that share common words and phrases, and group them together to form a smaller set of unique topics. We can express the topic similarity computation using the following formula:

\begin{equation} \label{eq1}
\begin{split}
TopicSimilarity(t_i, t_j) =\frac{t_{i} \cap t_{j}}{\min(t_i, t_j)}
\end{split}
\end{equation}


where \textit{t} represents a topic, which consists of a bag of topic-related words. Once we have calculated the topic similarity scores for all possible pairs of topics, we merge the topic pair that has the highest word similarity to form a new topic. We repeat this process iteratively until only $K$ unique topics remain, where $K$ represents the desired number of final topics.



\subsection{Topic Representation}


To evaluate the efficacy of our proposed PromptMTopic model, we assess the generated topics using standard topic model metrics, as detailed in Section 5. Evaluation necessitates that topics be represented as a word mixture. Given our model did not supply the required topic-word distributions for selecting the top W representative words, we computed the c-TF-IDF for each cluster and employed two methods to retrieve representative words. In the PromptMTopic-WSM approach, we select the top 10 words with the highest c-TF-IDF scores for each topic, under the assumption that a word's importance is closely tied to the topic's meaning. In the PromptMTopic-PBM approach, we prompt ChatGPT to return the top 10 words related to each topic, feeding it the top 100 words arranged in descending order of their c-TF-IDF scores. Table \ref{tab:topic_word_ordering_prompt} provides an example of the word ranking prompt using ChatGPT.

Using these two approaches, we can evaluate the effectiveness of our methodology and the coherence of the predicted topics.




%% file: Experiment.tex
\begin{table}[t!]
    \caption{Dataset statistics: Size represents the number of memes in each dataset, Caption represents average length of BLIP-2 caption per meme, Text represents average  length of superimposed text per meme.}
    {
    \begin{tabular}{cccc}
       \hline
       & & \multicolumn{2}{c}{\textbf{Avg. Length}} \\
        \cmidrule{3-4}
        \textbf{Dataset} & \textbf{Size} & \textbf{Caption} & \textbf{Text} \\
        \hline
        TotalDefMeme & 2,513 & 10.75 & 17.98 \\
        \hline
        FHM & 10,000 & 8.34 & 11.54 \\
        \hline
        Memotion & 6,992 & 8.44 & 14.84 \\
        \hline
    \end{tabular}
    }
    \label{tab:dataset_statistics}
\end{table}

\begin{table*}[ht]
  \small
  \centering
  \caption{Demonstration memes sampled from the FHM dataset, accompanied by their top five topic representation words generated by BERTopic and PromptMTopic-PBM. The words enclosed within the square brackets represent the topic names generated by PromptMTopic-PBM. We marked topic words related to the meme that do not accurately describe the meaning with \red{red}, and words that accurately describe the meaning with {\color{green}green}.}
  \begin{tabular}{p{2cm}|P{4.8cm}|P{4.8cm}|P{4.8cm}}
  \hline
  
    \multirow{2}{*}{\textbf{Memes}}
    
    & \begin{minipage}[!b]{0.58\columnwidth}
  \centering
  \raisebox{-.7\height}{\includegraphics[width=\linewidth]{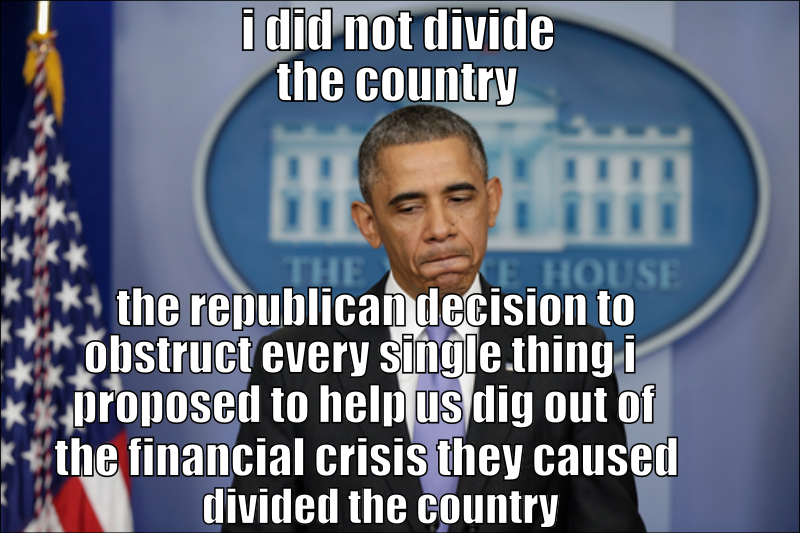}}
 \end{minipage}  
 &

    \begin{minipage}[!b]{0.58\columnwidth}
  \centering
  \raisebox{-.5\height}{\includegraphics[width=\linewidth]{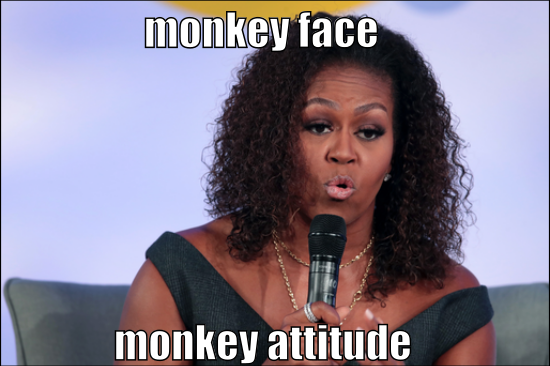}}
 \end{minipage} 
 &
    \begin{minipage}[!b]{0.54\columnwidth}
  \centering
  \raisebox{-.5\height}
  {\includegraphics[width=\linewidth]{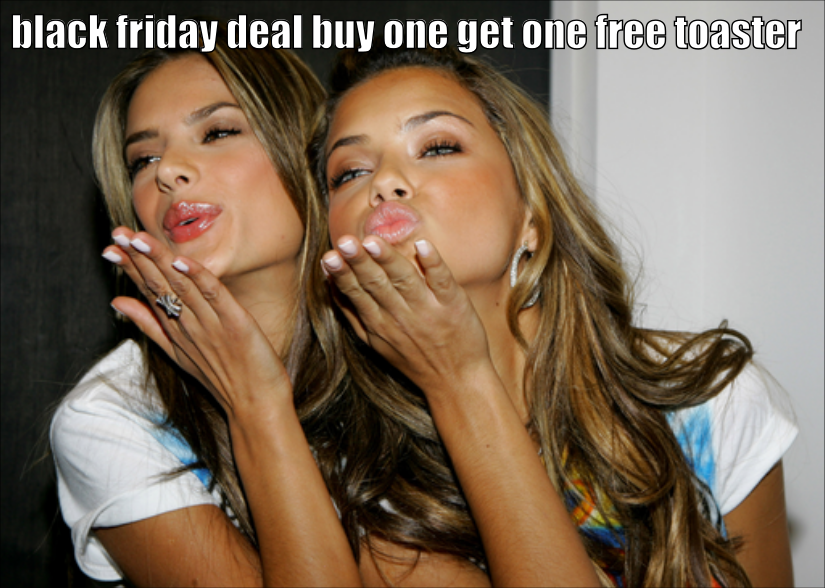}}
 \end{minipage}\\
    & (a) & (b) &  (c) \\
    \hline 
     \textbf{BERTopic}  &white, {\color{red}black}, muslim, girl, hair &  \red{obama}, \red{president}, {\color{red}michelle}, trump, donald & \red{shopping}, cart, walmart, \red{cleaner}, vacuum \\
    \hline

    \textbf{PromptMTopic-PBM} &
    
    \textbf{[politics]} trump, {\color{green}president}, democrats, {\color{green}republican}, election  &  \textbf{[politics]} trump, \red{president}, democrats, republican, election & \textbf{[finance]} bank, loan, income, financial, credit \\

    \cline{2-4}
    
    & \textbf{[finance]} bank, loan, income, {\color{green}financial}, credit & \textbf{[racism]} \textcolor{green}{racism}, {\color{green}race}, {\color{green}black}, white, people & \textbf{[shopping]} \red{shopping}, cart, store, \red{sale}, deal \\
    
    \hline
    \end{tabular}
  \label{tab:meme_demo}
\end{table*}

\begin{figure*}[t] 
	\centering
 	\includegraphics[scale = 0.52] {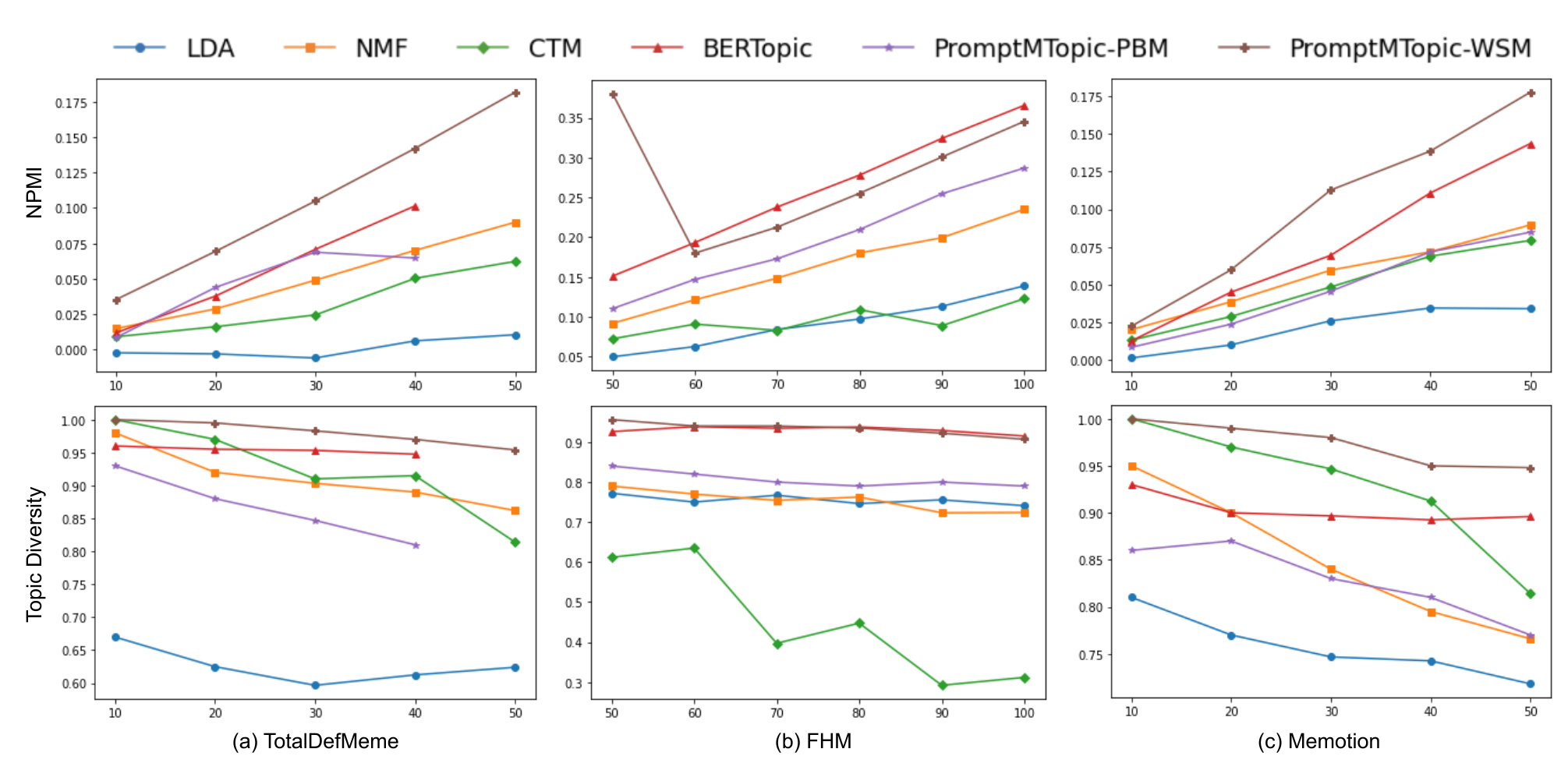} 
	\caption{NPMI and Topic Diversity plot for the three datasets for different $K$ values.}
	\label{fig:metrics_plot}
\end{figure*}

"In this section, we will discuss various multimodal datasets and baseline topic models. Then, we will present the results of experiments conducted to evaluate the performance of \textsf{PromptMTopic} in topic modeling for multimodal memes, compared to the baseline models. In addition, we will conduct a qualitative analysis to gain a deeper understanding of the topics generated by PromptMTopic.

\subsection{Experiment Settings}



\textbf{Datasets.} Our experimental analysis involves three publicly available meme datasets, each covering diverse contexts:

\begin{itemize}[leftmargin=*]
\item Facebook Hateful Memes (FHM) \cite{kiela2020hm}: FHM is centered around sensitive, prejudiced topics that pertain to individuals or groups based on inherent characteristics such as race, religion, and gender. The identification of these sensitive topics aids in effective intervention and prevention.
\item Total Defence Memes (TotalDefMeme) \cite{nirmal2023totaldefmeme}: TotalDefMeme features locally relevant topics related to Singapore's Total Defence concept, providing valuable insights into the associated events and activities.
\item Memotion \cite{ramamoorthy2022memo}: Memotion encapsulates viral, widespread topics on the internet, thereby facilitating a better understanding of current online trends.
\end{itemize}

Each of these datasets contains superimposed text annotated by human annotators. Table \ref{tab:dataset_statistics} presents the statistics for each dataset.

\textbf{Dataset Preprocessing.} To prepare the multimodal memes for baseline topic model training and ChatGPT prompting, we carried out a series of feature extraction steps detailed in Section \ref{subsection:feature_extraction}. This process often resulted in text and image features containing extraneous elements, such as meaningless words, website URLs, and usernames. The presence of these elements could negatively affect the performance of the baseline topic models. To rectify this, we added an extra preprocessing step, wherewe filtered out stop words, URLs, usernames, and punctuation from the text and image features using format matching. Then, we examined the top 30 most frequently occurring words, retaining meaningful terms and removing insignificant ones from both the text and image features. Finally, we concatenated the processed image and text features to form document features, which were then used to train the baseline topic models.

\textbf{Compared Models.} We will evaluate PromptMTopic against four well-established and widely used topic modeling models:
\begin{itemize}[leftmargin=*]
\item LDA \cite{blei2003latent}: LDA is a traditional generative probabilistic model that portrays documents as mixtures of topics. Each topic is characterized by a distribution over words.
\item NMF \cite{fevotte2011algorithms}: NMF is a non-probabilistic, decompositional model that factorizes a term-document matrix into two lower-ranking, non-negative term-topic and topic-document matrices.
\item CTM \cite{bianchi2020cross}: CTM is a neural topic model that leverages language-independent representations to formulate topic distributions.
\item BERTopic \cite{grootendorst2022bertopic}: BERTopic uses sentence embedding and clustering techniques to shape topics and employs a class-based variant of TF-IDF to generate topic representations. Unlike the other models, BERTopic assigns only one topic to each document.
\end{itemize}


In our evaluation of the topic models' effectiveness, we varied the number of topics ($K$) generated for each dataset. For the TotalDefMeme and Memotion datasets, we trained the baseline topic models starting with 10 topics, and increased the count in increments of 10 up to a maximum of 50. For the larger FHM dataset, we initiated with 50 topics and increased the count in increments of 10 up to a maximum of 100. It should be noted that, under the recommended configuration, BERTopic produces a maximum of 48 topics for the TotalDefMeme dataset. Similarly, when PromptMTopic-PBM is set to generate 50 topics for TotalDefMeme, some topics may end up with too few documents, resulting in topic representations of fewer than 10 words. Consequently, we have excluded the results for these models when $K$ equals 50 topics from the metrics plot.

\begin{table*}[!t]
\small
    \centering
    \caption{Qualitative evaluation of the topic-words representation. A subset of topics that contains the most number of documents are selected from PromptMTopics and its topics' name is used to manually find the most relevant topics generated by the baseline topic models. The related words belonging to the corresponding topic are highlighted in bold. }
    \begin{tabular}{l|ccc|ccc|ccc}
       \toprule
         \multirow{2}{*}{ \textbf{Models}}  & \multicolumn{3}{c}{\textbf{\textsc{TotalDefMeme}}} & \multicolumn{3}{c}{\textbf{\textsc{FHM}}} & \multicolumn{3}{c}{\textbf{\textsc{Memotion}}} \\
        & Topic 1 & Topic 2 & Topic 3 & Topic 1 & Topic 2 & Topic 3 & Topic 1 & Topic 2 & Topic 3 \\
       & politics & military & covid-19 & politics & racism & feminism & politics & relationships & technology \\
        \hline

        \multirow{5}{*}{\textbf{LDA}} 
        & \textbf{bmt} & \textbf{army} & \textbf{chinese} & \textbf{trump} & pants & \textbf{baby} & \textbf{obama} & iron & finding\\ 
 & cartoon & \textbf{military} & \textbf{covid} & \textbf{donald} & \textbf{white} & \textbf{dishwasher} & \textbf{president} & \textbf{girlfriend} & nemo\\ 
 
 & luke & \textbf{uniforms} & table & \textbf{president} & \textbf{dreadlocks} & black & \textbf{trump} & \textbf{person} & \textbf{technology}\\ 
 
 & phone & spiderman & indian & \textbf{wall} & vest & \textbf{kitchen} & \textbf{united} & said & dory \\ 
 
 & \textbf{military} & wrong & \textbf{19} & 
\textbf{trumps} & \textbf{nigger} & \textbf{washing} & \textbf{barack} & cowboy & fish\\  

         \hline

       \multirow{5}{*}{\textbf{NMF}} 
       & \textbf{pm} & \textbf{uniform} & \textbf{covid} & \textbf{obama} & \textbf{racist} & \textbf{girl} & \textbf{trump} & girl & \textbf{mark}\\ 
 & \textbf{lee} & \textbf{military} & \textbf{19} & \textbf{{michelle}} & \textbf{crime} & little & \textbf{donald} & \textbf{girlfriend} & \textbf{zuckerberg}\\ 
 & tie & \textbf{army} & \textbf{cases} & \textbf{president} & giving & jewish & \textbf{president} & little & \textbf{facebook}\\ 
 & pink & \textbf{soldier} & \textbf{ktv} & \textbf{barack} & \textbf{color} & \textbf{boy} & \textbf{trumps} & hey & \textbf{founder}\\ 
 & \textbf{hsien} & \textbf{uniforms} & \textbf{cluster} & \textbf{voters} & memes & pick & \textbf{wall} & boy & \textbf{congress}\\ 
        \hline


        \multirow{5}{*}{\textbf{CTM}} 
       & \textbf{lee} & \textbf{uniforms} & \textbf{covid} & \textbf{obama} & \textbf{white} & \textbf{girl} & \textbf{obama} & \textbf{friend} & \textbf{steve}\\ 
 & \textbf{minister} & \textbf{military} & \textbf{19} & \textbf{president} & \textbf{black} & little & \textbf{president} & text & \textbf{jobs}\\ 
 & \textbf{prime} & \textbf{ocs} & \textbf{cases} & \textbf{trump} & \textbf{racist} & looking & \textbf{house} & \textbf{friends} & harvey\\ 
 & show & \textbf{uniform} & airport & \textbf{president} & \textbf{jews} & \textbf{mom} & \textbf{barack} & \textbf{happy} & \textbf{computer}\\ 
 & \textbf{pm} & \textbf{bmt} & malaysia & \textbf{michelle} & \textbf{hitler} & old & first & \textbf{bad} & \textbf{technology}\\
        \hline

        \multirow{5}{*}{\textbf{BERTopic}} 
        & \textbf{chinese} & \textbf{gun} & \textbf{covid19} & \textbf{obama} & \textbf{racism} & \textbf{rights} & \textbf{obama} & minion & \textbf{zuckerberg}\\ 
 & \textbf{indian} & \textbf{uniform} & \textbf{vaccinated} & \textbf{president} & \textbf{nigger} & \textbf{womens} & \textbf{hillary} & minions & \textbf{mark} \\ 
 & \textbf{minister} & \textbf{military} & \textbf{covid} & \textbf{michelle} &  \textbf{africa} & \textbf{feminist} & \textbf{clinton} & wallpaper & \textbf{facebook}\\ 
 & \textbf{malay} & \textbf{police} & \textbf{cases} & \textbf{trump}  & \textbf{racist} & \textbf{cookbooks} & \textbf{president} & quotes & \textbf{founder}\\ 
 & \textbf{china} & \textbf{army} & fully & \textbf{donald} & tried & literature & \textbf{donald}& \textbf{love} & \textbf{facebooks}\\ 
        \hline

        \multirow{5}{*}{\pbox{2.0cm}{\textbf{PromptMTopic-}\\\textbf{WSM}}}
         & \textbf{protest} & cartoon & rights & dozen & \textbf{white} & bread & \textbf{trump} & baby & \textbf{zuckerberg}\\ 
         & ok & \textbf{ns} & \textbf{lockdown} & \textbf{ideology} & \textbf{black} & line & cat & \textbf{wife} & \textbf{mark}\\ 
         & \textbf{hong} & \textbf{police} & \textbf{quarantine} & \textbf{infidels} & \textbf{muslim} & waiting & \textbf{hillary} & \textbf{married} & \textbf{jobs}\\ 
         & \textbf{jailed} & group & \textbf{countries} & \textbf{pinko} & \textbf{muslims} & \textbf{feminist} & \textbf{white} & \textbf{good} & \textbf{steve}\\ 
         & \textbf{kong} & \textbf{saf} & hahaha & \textbf{liberals} & \textbf{trump} & \textbf{patriarchy} & \textbf{president} & \textbf{marrying} & \textbf{technology}\\ 
        \hline
   
        \multirow{5}{*}{\pbox{2.0cm}{\textbf{PromptMTopic-}\\\textbf{PBM}}}
         & \textbf{government} & \textbf{saf} & \textbf{mask} & \textbf{trump} & \textbf{racism} & \textbf{feminism} & \textbf{trump} & \textbf{girlfriend} & \textbf{computer}\\ 
         & \textbf{pm} & \textbf{bmt} & \textbf{cluster} & \textbf{president} & \textbf{race} & \textbf{feminist} & \textbf{clinton} & \textbf{love} & \textbf{facebook}\\ 
         & \textbf{minister} & \textbf{army} & \textbf{cases} & \textbf{democrats} & \textbf{black} & \textbf{women} & \textbf{putin} & \textbf{relationship} & \textbf{apple}\\ 
         & \textbf{pap} & \textbf{ns} & \textbf{virus} & \textbf{republican} & \textbf{white} & \textbf{abortion} & \textbf{obama} & \textbf{boyfriend} & \textbf{iphone}\\ 
         & \textbf{lee} & \textbf{soldiers} & \textbf{quarantine} & \textbf{election} & \textbf{people} & \textbf{rights} & \textbf{election} & \textbf{couple} & \textbf{robot}\\ 
        \hline
    \end{tabular}
    \label{tab:demo_table}
\end{table*}

\subsection{Topic Evaluation} 

\textbf{Quantitative Evaluation.} Evaluations of topic modeling often use two well-established metrics: topic coherence and topic diversity. Topic coherence gauges the extent to which the words within a topic are related, forming a coherent group. It is typically calculated using statistics and probabilities drawn from the reference corpus, focusing specifically on the context of the words. In our experiments, we employed Normalized Pointwise Mutual Information (NPMI) \cite{bouma2009tc} as our measure of topic coherence\footnote{We used 10e-6 for epsilon to prevent the logarithm of zero}, utilizing a Python implementation provided in a recent study on topic modeling metrics \cite{lim-lauw-2023-large}. A higher NPMI score signifies better coherence, with a perfect correlation being represented by a score of 1.

Conversely, topic diversity \cite{dieng2020td} evaluates the proportion of unique words across all topic representations. The diversity score ranges from 0 to 1, where a score of 0 indicates repetitive topics, and a score of 1 indicates diverse topics. This metric is crucial for ensuring that a topic model covers a wide range of themes without overemphasizing any particular topic. Using these two metrics together provides insights into the effectiveness of topic modeling algorithms in identifying both coherent and diverse topics.

Figure \ref{fig:metrics_plot} depicts the NPMI topic coherence and topic diversity scores of topics generated by the various models on the three datasets. We note that PromptMTopic-WSM consistently outperforms other baseline topic models on most datasets across both metrics. Conversely, PromptMTopic-PBM demonstrates performance comparable to some baseline topic models. The strong performance of PromptMTopic-WSM suggests that the use of ChatGPT in topic modeling can generate highly coherent and diverse topics.


\textbf{Qualitative Evaluation.}
Recent research has suggested that a low NPMI topic coherence score does not necessarily indicate poor topic quality, as the score often exhibits a weak correlation with human ratings \cite{hoyle2021automated}. To further evaluate the quality of the topics produced by the PromptMTopic model, we conducted a manual assessment of word coherence within the topics. This provides valuable human-centric insights into the quality of the generated topics. Specifically, we selected a subset of topics from the PromptMTopic-PBM model that contained the most documents. Using the names of these topics, we retrieved the most relevant topics generated by the PromptMTopic-WSM model and the baseline topic models. To ensure a fair comparison among the various models, we set the number of latent topics for each model based on the dataset: 40 topics for the TotalDefMeme and Memotion datasets, and 100 for the FHM dataset. This approach allowed us to carry out objective evaluations and comparisons of the various models.

Table \ref{tab:demo_table} presents the top five words per topic for each model across the three datasets. Despite the PromptMTopic-WSM model's higher NPMI topic coherence and topic diversity scores, our manual assessment surprisingly found the PromptMTopic-PBM model to provide better topic representation; all of its words can be linked to their corresponding topics. This indicates that using ChatGPT to select topic words can lead to more interpretable topics. This can be attributed to the extensive knowledge and strong language understanding encapsulated in large language models (LLMs), which enhance topic modeling and topic representation generation.

\subsection{Meme-Topic Analysis}

While both automatic and manual evaluations of the models' generated topics have been performed, the ability of these models to accurately understand and categorize memes remains uncertain. Therefore, we conducted a meme-topic analysis on the FHM dataset to evaluate the precision of topic assignment for the memes. Specifically, we scrutinized the BERTopic and PromptMTopic models to see if the related topic words within their assigned topics accurately represented the meaning of a given meme. Based on our evaluation, the BERTopic model demonstrated significant performance in quantitative metrics and produced topics mainly consisting of related topic words. On the other hand, the PromptMTopic-PBM model displayed its strength in the coherence of its generated topics, with all the topic words being closely related to the designated topic.

Table \ref{tab:meme_demo} displays three memes from our analysis, along with their corresponding topic words from the BERTopic and PromptMTopic-PBM models. We noted that the PromptMTopic-PBM model provided more suitable topics for certain memes. For instance, the BERTopic model associated the meme depicted in Table~\ref{tab:meme_demo}(a) with a topic that contained racial and gender terms. However, the meme did not contain any offensive or hateful remarks about racial or gender issues. Instead, it satirized the political divide between the Democratic and Republican parties. Therefore, although Barack Obama, being an African-American, could be associated with the topic word "black," the assigned topic did not accurately represent the meme's intended meaning. Conversely, the PromptMTopic-PBM model correctly assigned the topics of "politics" and "finance," which are more relevant to the meme's content.

Upon examining another meme in Table~\ref{tab:meme_demo}(b), we found that the BERTopic model categorized the meme under a topic containing politicians' names and the common name "Michelle." While Michelle Obama, as a former First Lady of the United States, can be linked with politicians, the meme primarily expresses racial hostility towards her as an African-American woman. Consequently, although the topic contains related terms, they do not convey the meme's intention. On the other hand, we noticed that the PromptMTopic-PBM model assigned both an inappropriate topic ("politics") and a more suitable topic ("racism") to the meme.

However, the PromptMTopic-PBM model, while better at assigning topics to memes, still struggled with accurately analyzing certain memes. Examining the meme in Table~\ref{tab:meme_demo}(c), we observe clear evidence of misogyny, with women treated as objects and means to household chores ("buy one get one free toaster"). However, both the PromptMTopic-PBM and BERTopic models failed to capture the meme's intended message and categorized it under inappropriate topics such as "shopping" and "finance." This suggests that the PromptMTopic and BERTopic models did not fully understand and interpret the memes, highlighting room for future improvements in multimodal meme topic modeling.

In conclusion, the proposed PromptMTopic model outperformed the baseline topic models. PromptMTopic-WSM achieved the highest NPMI topic coherence and topic diversity scores on most datasets, while PromptMTopic-PBM generated topics with highly related words and frequently assigned suitable topics to memes.

%% file: Conclusion.tex

Our paper introduces an innovative method for multimodal topic modeling on memes. Our approach outperforms well-established topic models across three distinct datasets. A qualitative analysis underscores our method's ability to discern the nuances and subtleties in meme content, thereby accurately identifying meaningful topics that existing models might overlook. The originality of our approach lies in the concurrent modeling of both meme modalities, leveraging the power of LLMs for topic modeling. We believe our findings extend beyond the realm of meme analysis. LLMs have demonstrated potential in a myriad of NLP tasks, and our study further illuminates their capability to enhance topic modeling across various domains. In future research, we plan to explore automated fine-tuning for LLMs for multimodal topic modeling.



%% file: Appendix.tex
\begin{figure*}[h]
	\centering
 	\includegraphics[width=\textwidth] {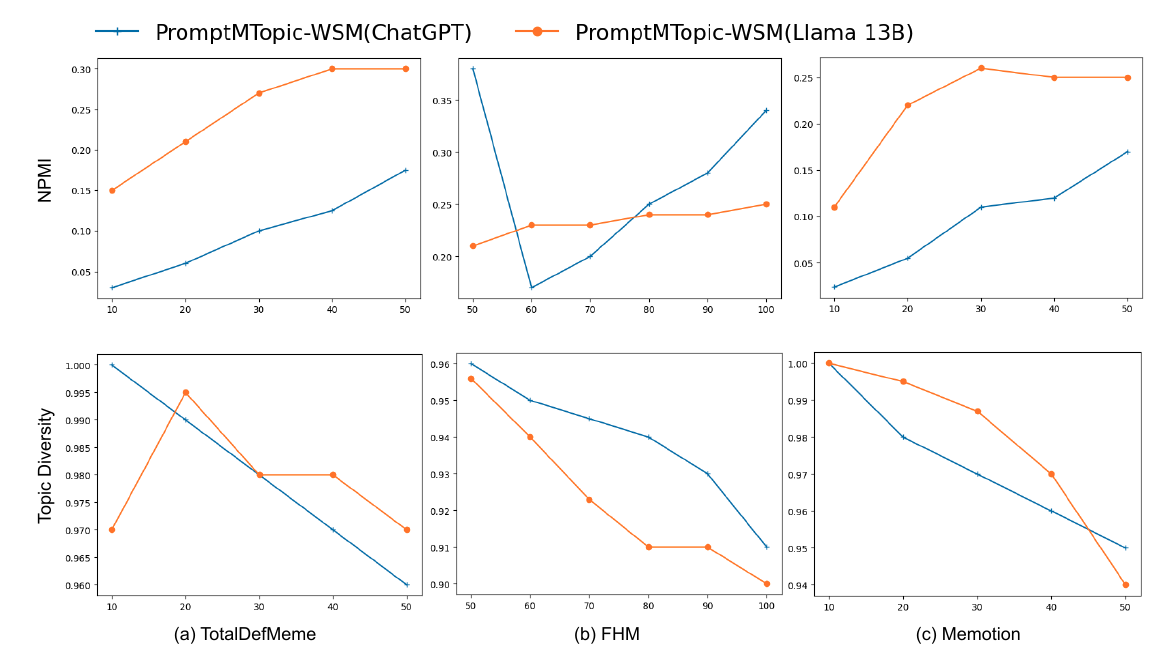} 
	\caption{NPMI and Topic Diversity plot for the three datasets for different k values.}
	\label{fig:AppendixA1}
\end{figure*}
\subsection{
Limitations
}
Although our evaluations demonstrate that large language models can be employed unsupervisedly to achieve superior topics, this approach bears its own limitations. In particular, LLMs are susceptible to hallucinations and intrinsic biases. While we've used demonstration examples to alleviate this issue, it nonetheless lingers. Additionally, our prompt design process has been manual, suggesting that exploring automatic prompt-tuning could potentially enhance performance. Moreover, the captions produced by BLIP-2 might not always encapsulate the essence of an image and can sometimes introduce unhelpful information to the model. Despite these limitations, we hope this work serves as a benchmark for assessing the potential of large language models in meme topic analysis and spurs further research to overcome these constraints.

\subsection{Topic Modeling using Llama13B\cite{touvron2023llama}}
We employed Llama13B model to extract topics, on the three datasets. We compare PromptMTopic-WSM using ChatGPT and Llama13B on coherence and diversity metrics. The results are shown in \ref{fig:AppendixA1} for different k values. We see that using Llama achieves a similar performance. This indicates the robustness of our proposed framework and signifies the potential of replacing LLMs in PromptMTopic with more sophisticated models in the future to improve performance even further. 

\subsection{Varying the number of demonstration examples}
we have conducted an experiment to analyze the impact of varying the number of demonstration examples when using ChatGPT and Llama13B. We utilized the TotalDefMeme dataset in this ablation study for the PromptMTopic-WSM framework, for k=20 topics. The number of demonstrations we tested ranged from 2 to 8, with increments of 2.
From table \ref{tab:appendixA2}, we observe that topic coherence and diversity remain comparable for different numbers of examples. 
\begin{table}[h]
    \caption{NPMI and Topic Diversity on TDefMeme dataset, using ChatGPT and Llama, k=20 topics, with different number of demonstration examples.}
    {
    \begin{tabular}{ccccc}
       \hline
        & \multicolumn{2}{c}{\textbf{NPMI}} & \multicolumn{2}{c}{\textbf{Topic Diversity}} \\
       
        \textbf{\#. Dems} & \textbf{ChatGPT} & \textbf{Llama13B}
        & \textbf{ChatGPT} & \textbf{Llama13B}\\

        \hline
        2 & 0.06 & 0.21 & 0.99 & 1.0\\
        \hline
        4 & 0.06 & 0.22 & 1.0 & 1.0\\
        \hline
        6 & 0.06 & 0.25 & 1.0 & 0.99\\
        \hline
        8 & 0.07 & 0.21 & 1.0 & 0.99\\
        \hline
    \end{tabular}
    }
    \label{tab:appendixA2}
\end{table}
\subsection{Distribution of unique and outlier topic clusters}
Examining the distribution of unique topic clusters in comparison to "outliers" (topics that didn't collapse into the existing list and were categorized as "Misc" for Prompt-based Matching) is indeed an important aspect of our analysis. Understanding how the model handles topic collapse while retaining coherence in light of dataset variance can provide essential insights into its robustness and applicability.

Interestingly, for PromptMTopic-PBM (prompt-based matching), we find that topic collapsing leads to a tiny fraction of topics, specifically between 1 to 1.6\% of the memes, being categorized under a 'Miscellaneous' topic. In PromptMTopic-WSM (word similarity matching), we do not have a ‘Miscellaneous’ topic in the merging process.

\subsection{Cost of PromptMTopic}
The expense associated with employing ChatGPT for the proposed approach is calculated at a rate of \$0.002 per 1000 tokens. Considering a meme dataset size of 10,000, the estimated cost of utilizing ChatGPT in all three stages of topic modeling, namely topic generation, topic collapsing, and topic ordering (akin to PromptMTopic-PBM), would amount to approximately \$20. Conversely, if we choose to employ ChatGPT solely for topic generation  (similar to \\PromptMTopic-WSM), the estimated cost would be approximately \$10.